\documentclass{esannV2}
\usepackage{graphicx}
\usepackage[latin1]{inputenc}
\usepackage{amssymb,amsmath,array}

\usepackage[usenames,dvipsnames]{xcolor}
\usepackage[colorlinks=true,urlcolor=teal,linkcolor=teal,citecolor=teal]{hyperref}
\usepackage{subcaption}
\DeclareMathOperator{\argmax}{arg\,max}

\begin{document}

\title{Why long model-based rollouts are no reason for bad Q-value estimates}

\author{Philipp Wissmann$^1$, Daniel Hein$^2$, Steffen Udluft$^2$, and Volker Tresp$^1$
%
\thanks{
The project this report is based on was supported with funds from the German Federal Ministry of Education and Research under project number 16ME0735K.
The sole responsibility for the report's contents lies with the authors.
}
%
\vspace{.3cm}\\
%
1- Ludwig-Maximilians-Universit\"at M\"unchen (LMU), Munich, Germany
%
\vspace{.1cm}\\
2- Siemens AG, Technology, Munich, Germany
}

\maketitle

\begin{abstract}
This paper explores the use of model-based offline reinforcement learning with long model rollouts.
While some literature criticizes this approach due to compounding errors, many practitioners have found success in real-world applications. 
The paper aims to demonstrate that long rollouts do not necessarily result in exponentially growing errors and can actually produce better Q-value estimates than model-free methods. 
These findings can potentially enhance reinforcement learning techniques.
\end{abstract}

\section{Introduction \& related work}
\label{section:intro}
While model-based reinforcement learning (RL) \cite{sutton1998introduction} is widely used by practitioners \cite{kober2013robotics,vop_paper2023Short}, it is often viewed critically due to theoretical considerations, \textit{e.g.}, \cite{talvitie2017self, janner2017whenShort}, especially in the case of long rollouts.
One of the arguments against model-based methods in RL literature is the worst-case error propagation of the one-step prediction error \cite{talvitie2017self}.
The \textit{hallucinated value hypothesis} by \cite{jafferjee2020hallucinating} contributes further to this by demonstrating that relying on model-predicted states can lead to misleading updates in planning tasks, see also \cite{abbas2020selectiveShort, chelu2020forethoughtShort}.
These arguments challenge the practitioners' positive perspective on model-based offline RL.
    
In this paper, we provide an explanation for why model-based RL can create effective policies even with long rollouts, although the model error for fixed action sequences increases exponentially with the number of steps. 
The reason for this is that most functioning algorithms optimize policies \cite{swazinna2019comparing}, not action sequences, and the policy is not \textit{blind} \cite{talvitie2017self} in the rollout, but is \textit{informed} and reacts to the respective situation simulated by the model in the rollout.

In the following, we demonstrate which drastic difference it makes for the modeling errors whether the policy is blind or whether it is informed and can react to the simulated state (Fig.~\ref{Fig:BlindRealTra}).
We compare the performance in estimating Q-values by model-free fitted Q evaluation (FQE), \textit{e.g.},~\cite{migliavacca2011fitted,hua2021bootstrappingShort}, with the one of model-based rollouts.
This comparison shows that rollout-based Q-value estimates can yield a significant lower estimation error.
Furthermore, to demonstrate the utility of the improved Q-value estimation, we modify the well established model-free Q-learning algorithm \textit{neural fitted Q iteration} (NFQ) \cite{riedmiller2005neuralnfqShort} by replacing the bootstrapping-based Q-value update with a bootstrapping-free rollout-based approach and show a significant gain in robustness during policy learning.

\section{Experimental setup}
\label{section:setup}
The experiments are performed using the cart-pole balancing benchmark.
The state space is four-dimensional, comprising the state variables position $x$, velocity $\dot{x}$, angle $\theta$, and angular velocity $\dot{\theta}$.
The data set $D$ has been generated by a random policy on the gym environment \textit{CartPole-v1} from the RL benchmark library \textit{Gymnasium}\footnote{\url{https://gymnasium.farama.org}}. 
$D$ consists of 20{\small,}000 observation tuples of form $(s_t,a_t,s_{t+1},r_t)$.
The environment terminates after an average of 22.3 steps because the pole falls over and leaves the permitted angular range, \textit{i.e.},~$|\theta| > 0.2095$. 
By using the random policy and initializing the cart near the center of the track (\textit{i.e.},~$x \approx 0$) and the pole near the upright position (\textit{i.e.},~$\theta \approx 0$), no data is generated near the boundaries of the track ($|x|=2.4$), particularly not with the pole upright far from the center of the track. 

For the reward, we define a function that assigns 1 for an upright pole with the cart in the center and decreases quadratically along cart position and pole angle relative to their termination bounds, \textit{i.e.}, $r = (1 - (x/2.4)^2 + 1 - (\theta/0.2095)^2)/2$.

\subsection{Models}
The transition model $M$ comprises four sub-models, one for each of the four state variables ($x$, $\dot{x}$, $\theta$, $\dot{\theta}$). Each sub-model is a feed forward neural network (NNs) with a 5-16-1 architecture and ReLU nonlinearity, uses the four state variables and the action as input, and fits for its respective state variable the differences between the next and current state.
For the reward model $R$, a feed forward NN with a 9-16-1 architecture is trained using (state-action-next state) as input to predict the reward.

The data set is split with a 70:30 ratio into a training and a validation set.
We use the Adam algorithm with a learning rate of 0.01 and mini-batch updates with 100 samples from the training set.
An early stopping strategy halts the training if no improvement of the validation error is made in 100 epochs and the best parameters found so far are persisted. 
    
In order to investigate the effect of the model's precision, we experiment with models of different qualities by also stopping the training after one epoch, ten epochs, and one hundred epochs.
Fig.~\ref{fig:learning_curve_pole_angle} shows exemplary the training process for the pole angle model. 

\section{Blind vs. informed policy rollouts}

For a given start state $s_t\in S$ and a given policy $\pi$, the transition model $M$ can be used to generate a trajectory called a rollout, where $\tilde{s}_{k+1}={M}(\tilde{s}_k,a_k)$. 
$a_k$ is either defined by an action sequence for the blind policy, or by $a_k=\pi(\tilde{s}_k)$ for the informed policy. This is repeated for $K$ steps, where $K$ is the rollout length.

\begin{figure}[ht!]
    \begin{subfigure}{0.245\textwidth}
        \includegraphics[width=1\textwidth]{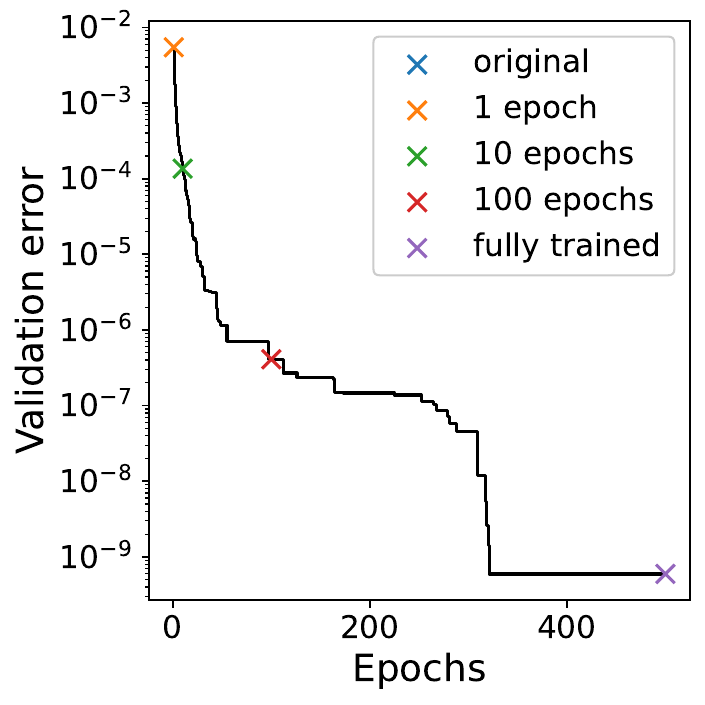}
        \subcaption{Pole angle model}
        \label{fig:learning_curve_pole_angle}
    \end{subfigure}
    \begin{subfigure}{0.75\textwidth}
        \includegraphics[width=1\textwidth]{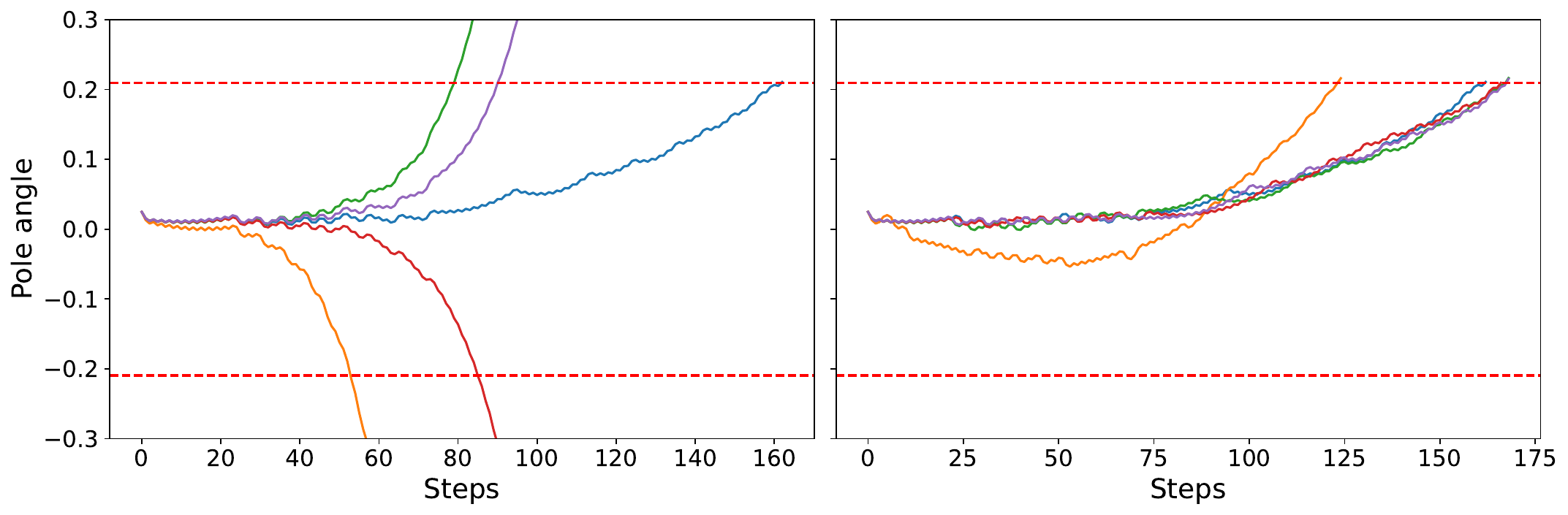}
        \subcaption{Blind vs. informed rollouts on transition models}
        \label{fig:blind_realistic_rollouts}
    \end{subfigure}
    \caption{Comparing the effect of different model qualities.
    (a) Learning curve of a pole angle model. 
    Crosses highlight the epochs in which the weights have been saved.
    (b) Difference between predicting a state trajectory through blind (left) and informed (right) policy rollout.
    }
    \label{Fig:BlindRealTra}
\end{figure}

As can be seen in Fig.~\ref{fig:blind_realistic_rollouts}, the rollouts in the case of the blind policy deviate progressively from the true trajectory as the number of steps increases. 
It is also evident that the better the model, the longer the deviation remains small. However, due to the accumulating errors, no model is good enough to stay close to the true trajectory over hundreds of steps.
The plot on the right side of Fig.~\ref{fig:blind_realistic_rollouts} shows a completely different picture. In the case of informed policies, which receive the simulated state as input and can react accordingly to the simulated state, the discrepancy to the true trajectory remains small for all but the model trained for only one epoch.
It is also noteworthy that even the model that was only trained for ten epochs and has a significantly higher one-step error than the fully trained model (see Fig.~\ref{fig:learning_curve_pole_angle}) only deviates slightly from the true trajectory in the rollout.

\section{Q-value estimates for policies}
\label{section:PolEval}

As shown above, model rollouts of an informed policy can be quite similar to the true trajectory. 
In order to substantiate this statement quantitatively, model rollouts will now be used to estimate the state-value $\tilde{V}^\pi$ of an informed policy $\pi$ and subsequently compared with the true return measured on the gym environment.

Since we want to study long rollouts, we set the discount factor $\gamma$ close to 1, \textit{i.e.}, $\gamma = 0.99$ and $K = 1,000$.
The estimated model-based state-value for deterministic policy $\pi$ starting from state $s_t\in S$ 
is computed by:
\begin{equation}
    {\tilde{V}}^{\pi}_{\text{MB}}(s_t) = \sum^{K-1}_{k=0} \gamma^{k} R(\tilde{s}_{k},\pi(\tilde{s}_{k}),\tilde{s}_{k+1}),
    \label{eq:rollout_return}
\end{equation}
where the rollout start is set to $\tilde{s}_{0}=s_t$, and $\tilde{s}_{k+1}=M(\tilde{s}_k,\pi(\tilde{s}_{k}))$.

To compare the quality of the rollout-predicted state-values with the ones of a model-free method, we used the well-known FQE algorithm with an NN with architecture 5-64-1 and ReLU activation as Q-function:
\begin{equation}
    Q_{i+1}(s_t,a_t)\leftarrow r_t+\gamma Q_i(s_{t+1},\pi(s_{t+1})).
    \label{eq:fqe}
\end{equation}
Note that FQE is an algorithm which iteratively builds up a Q-function by bootstrapping from its own Q-function of the previous iteration.
The model-free FQE estimated state-value of policy $\pi$ can be computed by: 
\begin{equation}
    \tilde{V}^{\pi}_{\text{MF}}(s_t)=Q(s_t,\pi(s_t)).
\end{equation}

\begin{figure}[ht!]
    \begin{subfigure}{0.49\textwidth}
        \includegraphics[width=1\textwidth]{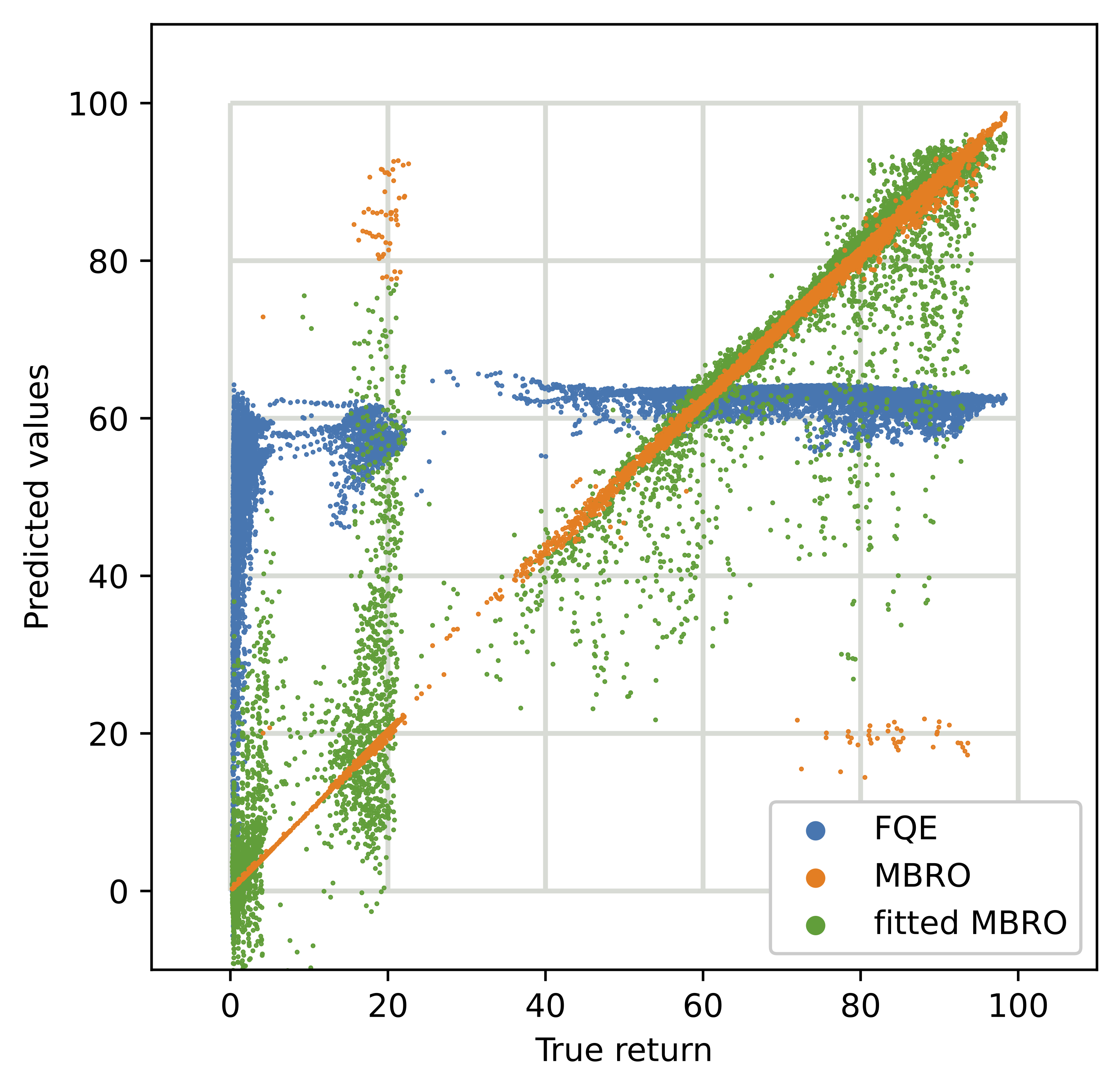}
        \subcaption{Predicted state-values}
        \label{fig:scatter_q}
    \end{subfigure}
    \begin{subfigure}{0.5\textwidth}
        \includegraphics[width=1\textwidth]{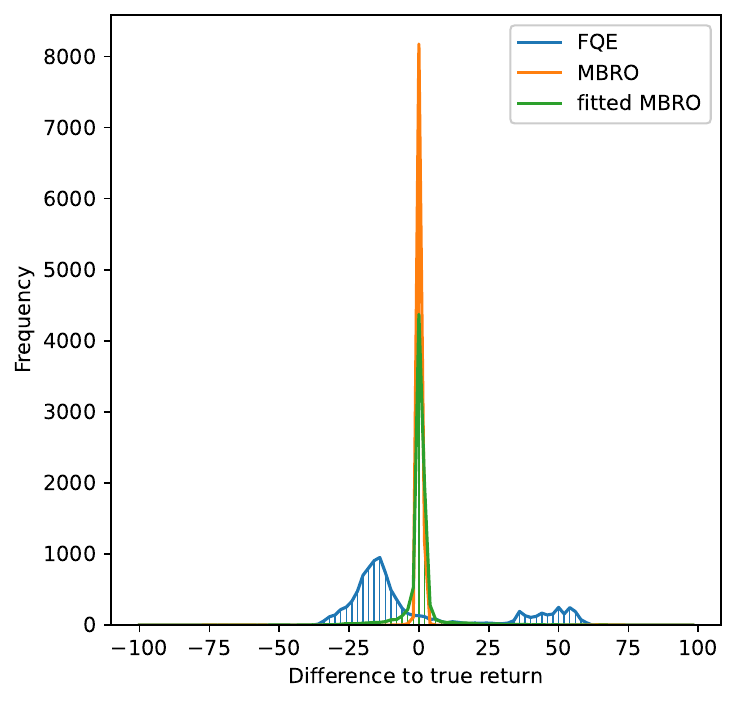}
        \subcaption{Error of predicted state-values}
        \label{fig:histogram_q}
    \end{subfigure}
    \caption{Comparison of predicted state-values.
        (a) Scatter plot of predicting state-values model-free (FQE), with model-based rollouts (MBRO), and fitted MBRO. 
        (b) Histogram of the differences of predicted state-values and the true return.
        }\label{Fig:ScatterHist}
\end{figure}

Fig.~\ref{Fig:ScatterHist} shows huge quality differences comparing the model-based $\tilde{V}^{\pi}_{\text{MB}}$ with the model-free $\tilde{V}^{\pi}_{\text{MF}}$.
To verify that the performance difference does not stem from a poorly chosen NN architecture of the Q-function, we fitted an NN of the same layout using the model-based state-value estimates and evaluated it.
Results in Table~\ref{Tab:RMSEforPolicyEvaluation} show that the used NN layout is capable of yielding good state-value estimates.

\begin{table}[ht!]
  \centering
  \begin{tabular}{|c||c|c|c|}
    \hline
    & FQE & MBRO & fitted MBRO \\
    \hline
    QNN architecture & 5-64-1 & N/A & 5-64-1 \\
    \hline
    Q-value RMSE & $26.1 \pm 0.2$ & $\textbf{3.9} \pm \textbf{0.2}$ & $6.36 \pm 0.04$ \\
    \hline
    Correlation coefficient & $0.699 \pm 0.004$ & $\textbf{0.9923} \pm \textbf{0.0009}$ & $0.9793 \pm 0.0002$ \\
    \hline
  \end{tabular}
  \caption{Errors and correlations of predicted state-values for different algorithms: 
    Model-free (FQE) and model-based (MBRO and fitted MBRO).
    Shown are averages over multiple seeds with their standard errors. 
    }\label{Tab:RMSEforPolicyEvaluation}
\end{table}

\section{Policy learning}
\label{section:PolLearning}

In the previous section, we demonstrated that model-based state-value estimations can be significantly better compared to model-free estimations in offline policy evaluation.
Next, we will investigate whether Q-value-based offline RL methods can benefit from this observation.

\begin{figure}[ht!]
    \begin{subfigure}{0.62\textwidth}
        \includegraphics[width=1\textwidth]{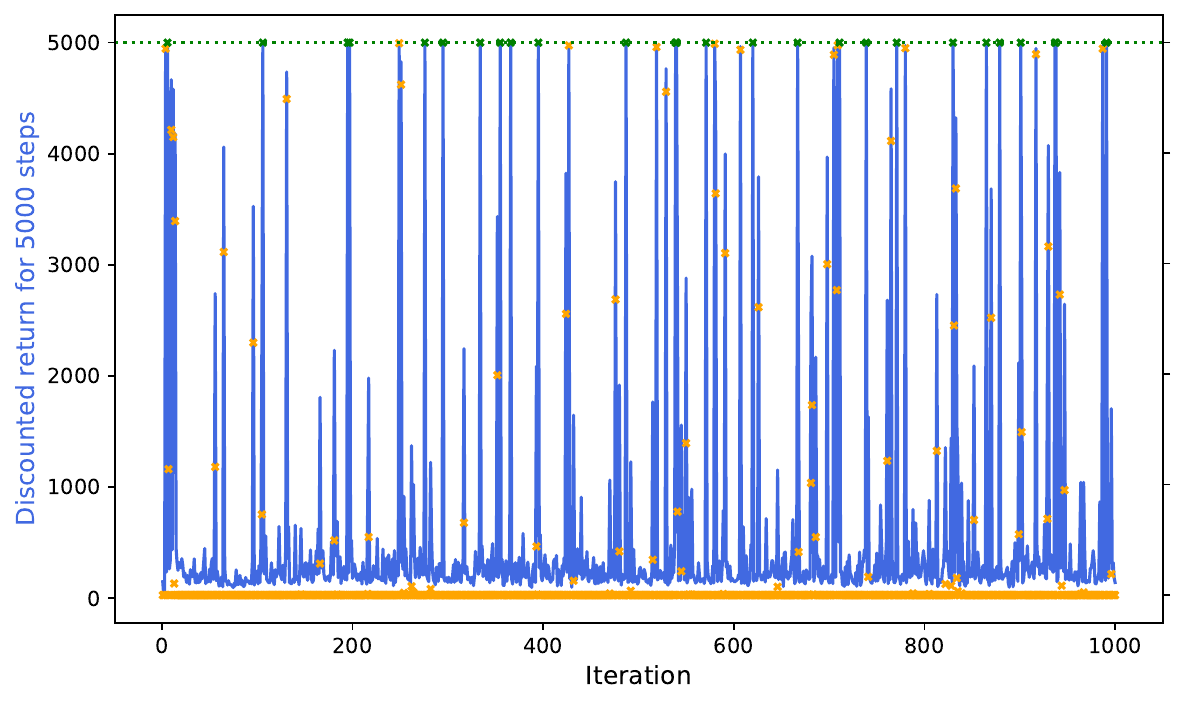}
        \subcaption{Performance of NFQ}
        \label{fig:policy_nfq}
    \end{subfigure}
    \begin{subfigure}{0.37\textwidth}
        \includegraphics[width=1\textwidth]{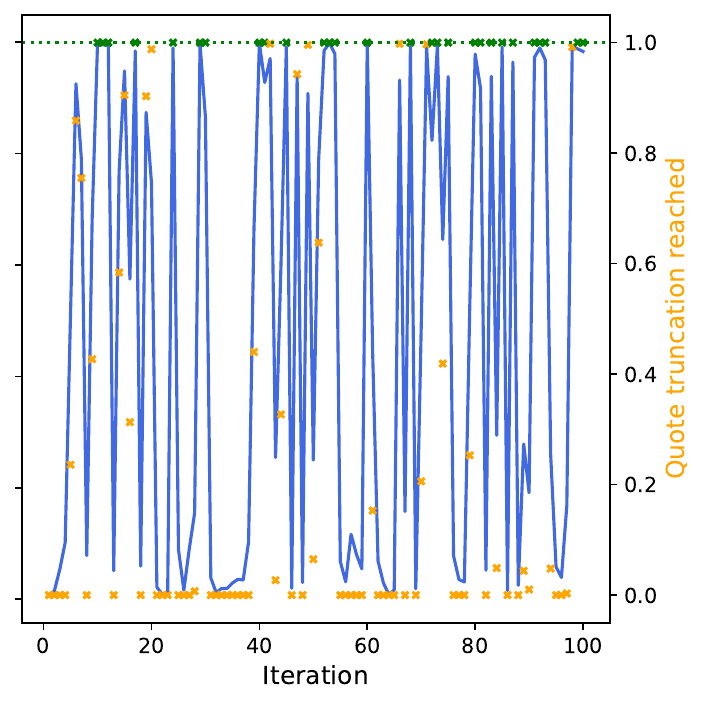}
        \subcaption{Performance of BSF-NFQ}
        \label{fig:policy_bsf}
    \end{subfigure}
    \caption{Iteration-wise policy performance averaged over 1,000 gym environment episodes.
        Blue lines represent the average discounted return over 1{\small,}000 episodes each with 5{\small,}000 steps.
        Cross markers depict the quote of episodes reaching 5{\small,}000 steps.
        Green markers represent iterations where \textit{perfect} policies have been found, \textit{i.e.}, policies balanced successfully in all episodes for at least 5{\small,}000 steps.
        }\label{Fig:PolicyPerf}
\end{figure}

NFQ \cite{riedmiller2005neuralnfqShort} is a well-known model-free offline RL method which learns an optimal policy iteratively by modifying FQE Eq.~\ref{eq:fqe} in the following way:
\begin{equation}
    Q_{i+1}(s_t,a_t) \leftarrow r_t+\gamma \max_{a_{t+1}} Q_i(s_{t+1},a_{t+1}).
    \label{eq:nfq}
\end{equation}
In each iteration $i$, the maximum Q-value of the previous Q-function for state $s_{t+1}$ is used to compute the new targets, which yields an optimal policy given by $a_t = \argmax_a Q(s_t,a)$.
Fig.~\ref{fig:policy_nfq} depicts a typical NFQ learning run over 1{\small,}000 iterations.
Note that the learning process of NFQ is rather unstable and it is only successful in $3.8\%$ of the iterations.

To test whether model-based rollout state-value estimates can improve NFQ's performance on our benchmark, we replaced the bootstrapping-based Q-value estimation with the rollout estimation from Eq.~\ref{eq:rollout_return}:
\begin{equation}
    Q_{i+1}(s_t,a_t) \leftarrow r_t+ \gamma {\tilde{V}}^{\pi}_{\text{MB}}(s_{t+1}), \text{ with } \pi(s)=\argmax_a Q_i(s,a).
    \label{eq:bsf}
\end{equation}

Fig.~\ref{fig:policy_bsf} depicts the learning performance of this RL method called boot\-strap\-ping-free NFQ (BSF-NFQ).
Since BSF-NFQ does not need to build up the Q-values iteratively like NFQ, and the calculation of ${\tilde{V}}^{\pi}_{\text{MB}}(s_{t+1})$ makes the algorithm considerably slower, we performed only 100 iterations.
However, in these 100 iterations BSF-NFQ yielded on average 
 in $23.3\%$ of the iterations 
optimal policies.
Replacing the bootstrapping in NFQ by model-based rollout state-value estimates dramatically improved the robustness of the learning algorithm (see Table~\ref{Tab:PolicyLearningResults}).

\begin{table}[ht!]
  \centering
  \begin{tabular}{|c||c|c|}
    \hline
    & NFQ & BSF-NFQ \\
    \hline 
    Learning iterations & 1{\small,}000 & 100 \\
    \hline   
    Ratio of optimal policies [\%]& $3.8 \pm 0.4$ & $\textbf{23.3} \pm \textbf{1.1}$ \\
    \hline
  \end{tabular}
  \caption{Comparing learning results of NFQ with BSF-NFQ. 
  Each experiment (Fig.~\ref{Fig:PolicyPerf}) has been repeated ten times with different random seeds.}
  \label{Tab:PolicyLearningResults}
\end{table}

\section{Conclusion}

In this paper, we demonstrated that long rollouts in model-based RL do not always lead to exponentially growing errors. 
We have shown the drastic difference it makes for the modeling errors whether the policy in the rollout is blind or whether it is informed and can react to the simulated state.
We were able to show that long rollouts with informed policies can indeed provide better Q-value estimates compared to model-free methods and that using such Q-value estimates instead of bootstrapping can increase the robustness of policy learning.


\begin{footnotesize}

\bibliographystyle{unsrt}
\bibliography{references.bib}

\end{footnotesize}


 \end{document}